\documentclass{article}
\usepackage{spconf,amsmath,graphicx,hyperref}
\usepackage{cleveref} 
\usepackage{booktabs}
\usepackage{multicol}
\usepackage{multirow}
\usepackage[table]{xcolor}
\usepackage{amssymb}

\title{Hybrid Gaussians for Robust Open-Vocabulary 3D Segmentation with Multi-View Object Association and Boundary Refinement}
\name{
Xueqi Qiu$^{1}$,
Yueming Sun$^{1}$,
Tianyu Zhang$^{1}$,
Yuxuan Xia$^{2}$,
Yang Long$^{1}$%
}

\address{
$^{1}$Department of Computer Science, Durham University, Durham, United Kingdom\\
$^{2}$Shanghai Jiao Tong University, Shanghai, China
}
\begin{document}
%
\maketitle
\begin{abstract}
Open-vocabulary 3D segmentation localizes objects from free-form text queries, but remains challenging in real image sequences: incomplete or noisy 2D supervision destabilizes multi-view identity assignment, while full-scene semantic learning weakens object-level discriminability. We introduce Hybrid Gaussians, a unified 3D representation jointly modeling object association and language-aligned semantics. Its Multi-View Object Association mechanism combines Observation Fusion and Semantic Contrastive Learning to improve identity consistency and semantic discrimination. Boundary Reconstruction Optimization further refines local boundary structure to improve contour quality. Experiments on LERF and 3D-OVS demonstrate strong quantitative and qualitative performance. Our method achieves 59.1\% mIoU on LERF, yielding a 13.4\% relative gain over the baseline. Project page: \url{https://nora202.github.io/hybridgaussians}.

\end{abstract}
\begin{keywords}
Open-Vocabulary Segmentation, 3D Gaussian Splatting, Object-Aware Supervision
\end{keywords}
\section{Introduction}
\label{sec:intro}

Open-vocabulary 3D segmentation localizes and segments objects in 3D scenes using free-form text queries rather than predefined labels. It supports embodied perception, robotic interaction, immersive media, and language-guided scene understanding \cite{yan2025dynamic,he2026survey}. Existing methods lift supervision from 2D foundation models into point-based 3D representations \cite{peng2023openscene,takmaz2023openmask3d,yamazaki2024open}, while NeRF \cite{mildenhall2021nerf} and 3D Gaussian Splatting (3DGS) \cite{kerbl20233d} provide rendering-friendly alternatives. LERF, 3D-OVS, Feature-3DGS, LangSplat, and LEGaussians incorporate language-aligned features for text-driven querying and segmentation \cite{kerr2023lerf,liu2023weakly,zhou2024feature,qin2024langsplat,shi2024language}. Object-level modeling has also been explored through identity-aware grouping, object-specific Gaussian sets, and object-centric anchors \cite{ye2024gaussian,lu2025segment,zhu2025objectgs}. More recent methods combine instance and semantic cues through instance-aware grouping, identity-guided feature learning, cross-modal semantic alignment, instance-to-language mapping, or multi-view semantic aggregation \cite{wu2024opengaussian,jang2025identity,li2025vaf,cen2025laga,zhu2025cos3d}. Nevertheless, incomplete, noisy, or intermittent 2D cues still hinder consistent segmentation across views.

\begin{figure*}[t]
    \centering
    \includegraphics[width=\textwidth]{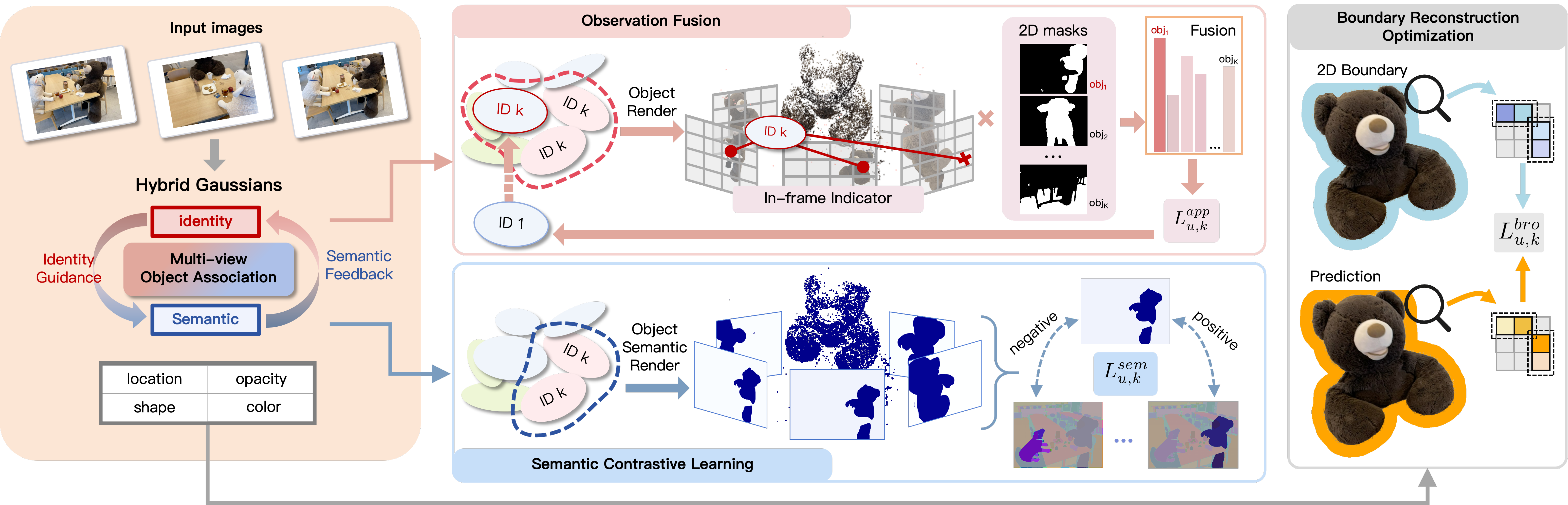}
    \vspace{-0.6cm}
    \caption{\textbf{Training pipeline of our method.} We jointly optimize Hybrid Gaussians through Multi-view Observation Fusion for identity reassignment, Semantic Contrastive Learning for object-aware semantic discrimination, and Boundary Reconstruction Optimization for sharper object boundaries, enabling robust open-vocabulary 3D segmentation.}
    \vspace{-0.3cm}
    \label{fig:training}
\end{figure*}

We therefore introduce Hybrid Gaussians, a unified 3D representation that equips each Gaussian with an object identity and a learnable language-aligned semantic embedding. Unlike prior methods that primarily use identity or semantic features as auxiliary attributes, our representation jointly exploits them for object-conditioned rendering and supervision. Building on this representation, Multi-view Object Association comprises two complementary components. Observation Fusion aggregates object-level evidence across views to update Gaussian identities, while Semantic Contrastive Learning combines object-conditioned semantic rendering with cross-view contrastive alignment to improve semantic discrimination. We further introduce Boundary Reconstruction Optimization, which aligns rendered gradients with masked image observations near object contours to improve boundary quality. In summary, our contributions are:

\begin{itemize}
\item We introduce a hybrid Gaussian formulation that jointly maintains object identities and language-aligned semantic embeddings within the same primitives, enabling object-conditioned rendering and learning.

\item We address inconsistent cross-view object association and weak object-level semantic discrimination through Multi-view Object Association.

\item We improve contour quality using a lightweight Boundary Reconstruction Optimization objective.

\item Experiments on LERF and 3D-OVS demonstrate strong qualitative and quantitative performance.
\end{itemize}

\section{Method}

\subsection{Hybrid Gaussians}

Given multi-view images
$\mathcal{I}=\{I^{gt}_{u}\}_{u\in\mathcal{U}}$, we represent the scene using $N$ Gaussian primitives following 3DGS \cite{kerbl20233d}:
\begin{equation}
\mathcal{G}
=
\left\{
g_n=
(l_n,\Sigma_n,\delta_n,c_n,z_n,f_n)
\right\}_{n=1}^{N},
\label{eq:hybrid_gaussians}
\end{equation}
where $l_n\in\mathbb{R}^{3}$, $\Sigma_n\in\mathbb{R}^{3\times3}$, $\delta_n$, and $c_n\in\mathbb{R}^{3}$ denote the center, covariance, opacity, and color, respectively. We augment each Gaussian with an object assignment $z_n\in\{0,\ldots,K\}$ and a learnable language-aligned embedding $f_n\in\mathbb{R}^{D}$. These identity and semantic attributes form Hybrid Gaussians for object-conditioned rendering and learning.

\subsection{Multi-view Object Association}
\label{section:moa}

\noindent\textbf{Observation Fusion (OF).}
We consider $K$ foreground objects and a background label indexed by $k\in\{0,\ldots,K\}$. SAM \cite{kirillov2023segment} initializes the object masks, which are propagated across views using SAM2 \cite{ravi2025sam}. After confidence filtering, the foreground mask of object $k$ in view $u$ is
\begin{equation}
M_{u,k}
=
\mathbb{I}\!\left[\zeta(I^{gt}_{u},k)\geq\tau_1\right],
\qquad k\in\{1,\ldots,K\},
\end{equation}
and the background is defined as
$M_{u,0}=1-\bigvee_{k=1}^{K}M_{u,k}$.

For Gaussian $g_n$ with center $l_n$, its projected location in view $u$ is obtained as
\begin{equation}
\textstyle
\tilde p_{n,u}=K_u(R_ul_n+t_u),
\qquad
p_{n,u}
=
\left(
\frac{\tilde p^x_{n,u}}{\tilde p^z_{n,u}},
\frac{\tilde p^y_{n,u}}{\tilde p^z_{n,u}}
\right).
\end{equation}
We define $V_{n,u}\in\{0,1\}$ to indicate whether $g_n$ has positive depth and its projected center lies within the image domain. The multi-view support for assigning $g_n$ to object $k$ is then
\begin{equation}
s_{n,k}
=
\frac{
\sum_{u\in\mathcal U}
w_u V_{n,u}M_{u,k}(p_{n,u})
}{
\sum_{u\in\mathcal U}
w_uV_{n,u}+\epsilon
},
\label{eq:fused_observation}
\end{equation}
where $w_u$ denotes the learnable normalized reliability of view $u$.
For Gaussians with valid observations, the fused evidence is converted into a soft object association with temperature $T$:
\begin{equation}
q_{n,k}
=
\frac{\exp(s_{n,k}/T)}
{\sum_{j=0}^{K}\exp(s_{n,j}/T)}.
\label{eq:identity_update}
\end{equation}
The hard identity $z_n$ is assigned to the object with the highest association probability. For Gaussians without valid observations, both the soft association and hard identity are retained from the preceding iteration.

Unlike single-view hard assignment, the soft association aggregates consistent object evidence across views and reduces the influence of missing, occluded, or locally ambiguous masks. The association probability directly modulates the contribution of $g_n$ when rendering object $k$:
\begin{equation}
\textstyle
I^{pred}_{u,k}(p)
=
\sum_{n=1}^{N}
\left\{
q_{n,k}c_n\alpha_{n,u}(p)
\prod_{m<n}
\left(1-\alpha_{m,u}(p)\right)
\right\}.
\label{eq:soft_object_rendering}
\end{equation}
The object-conditioned rendering is supervised using
\begin{equation}
\textstyle
\mathcal{L}^{app}_{u,k}
=(1-\lambda)\|I^{pred}_{u,k}-I^{gt}_{u,k}\|_1
+\lambda[1-\operatorname{SSIM}(I^{pred}_{u,k},I^{gt}_{u,k})],
\label{eq:appearance_loss}
\end{equation}
where $I^{gt}_{u,k}=I^{gt}_{u}\odot M_{u,k}$.
This rendering objective couples multi-view object association with appearance reconstruction, allowing the soft associations to be progressively refined during optimization and the corresponding hard identities \(z_n\) to be updated accordingly.

\noindent\textbf{Semantic Contrastive Learning (SCL).}
Although Observation Fusion improves object identity consistency, directly lifting dense 2D semantic features into 3D may still introduce cross-object feature contamination
\cite{wu2024opengaussian}. We therefore use the fused object
associations to perform object-conditioned semantic contrastive
learning.

For view $u$, we extract a dense language-aligned reference map from the spatial features of a frozen CLIP image encoder
$\Phi_i(\cdot)$ \cite{radford2021learning}:
\begin{equation}
S^{ref}_{u}
=
\operatorname{Up}\!\left(
\Phi_i^{\mathrm{dense}}(I^{gt}_{u})
\right)
\in\mathbb{R}^{D\times H\times W},
\label{eq:semantic_reference}
\end{equation}
$\operatorname{Up}(\cdot)$ upsamples the patch-level features to the image resolution. Using the soft object association $q_{n,k}$ obtained from Observation Fusion, we render an object-conditioned semantic map:
\begin{equation}
\textstyle
S^{pred}_{u,k}(p)
=
\sum_{n=1}^{N}
\left\{
q_{n,k}f_n\alpha_{n,u}(p)
\prod_{m<n}
\left(1-\alpha_{m,u}(p)\right)
\right\}.
\label{eq:fpred_obj}
\end{equation}
where the semantic embeddings are composited using the standard 3DGS alpha-blending weights.

To remove view-dependent spatial layouts, we aggregate the semantic features within the propagated object mask:
\begin{equation}
\textstyle
z^{ref}_{u,k}
=
\frac{
\sum_p M_{u,k}(p)S^{ref}_{u}(p)
}{
\sum_p M_{u,k}(p)+\epsilon
},
\,
z^{pred}_{u,k}
=
\frac{
\sum_p M_{u,k}(p)S^{pred}_{u,k}(p)
}{
\sum_p M_{u,k}(p)+\epsilon
}.
\label{eq:object_pooling}
\end{equation}
We further apply $\ell_2$ normalization to obtain
$\bar z^{ref}_{u,k}$ and $\bar z^{pred}_{u,k}$. For an anchor $\bar z^{pred}_{u,k}$, the reference embedding of the
same object in another valid view $v\neq u$ forms a positive pair, whereas the embeddings of other foreground objects form negatives. The semantic contrastive objective is
\begin{equation}
\mathcal{L}^{sem}_{u,k}
=
-\log
\frac{
\exp\!\left(
\operatorname{sim}
(\bar z^{pred}_{u,k},\bar z^{ref}_{v,k})/\tau_2
\right)
}{
\displaystyle
\sum_{k'=1}^{K}
\exp\!\left(
\operatorname{sim}
(\bar z^{pred}_{u,k},\bar z^{ref}_{v,k'})/\tau_2
\right)
},
\label{eq:infonce_sem}
\end{equation}
$\operatorname{sim}(\cdot,\cdot)$ denotes cosine similarity and
$\tau_2$ is the temperature.

Observation Fusion maintains consistent object identities for Gaussians across views, while Semantic Contrastive Learning improves intra-object consistency and inter-object separability. Together, they form Multi-view Object Association, enabling stable object identities and discriminative language-aligned representations.

\begin{figure}[t]
    \centering
    \includegraphics[width=\linewidth]{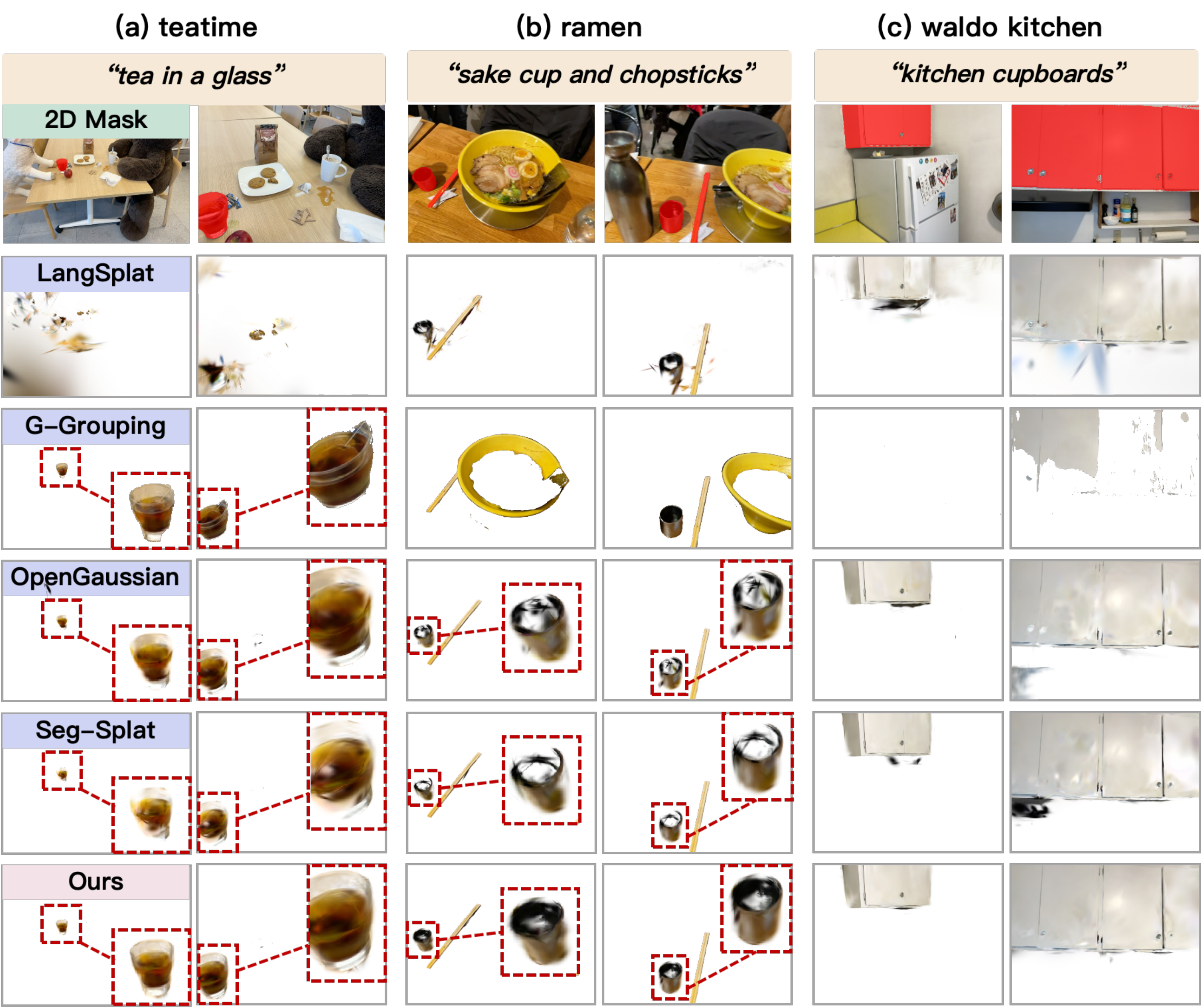}
    \vspace{-0.4cm}
    \caption{\textbf{Qualitative comparison.} For each text query, we visualize the corresponding object segmentation results from different views. Gaussian Grouping \cite{ye2024gaussian} and Segment-then-Splat \cite{lu2025segment} are abbreviated as G-Grouping and Seg-Splat.}
    \vspace{-0.2cm}
\label{fig:qualitative_results}
\end{figure}

\begin{table}[t]
\centering
\caption{\textbf{Quantitative comparison on LERF and 3D-OVS in terms of average mIoU (\%).} The best and second-best results are highlighted in bold and underlined, respectively.}
\resizebox{0.6\linewidth}{!}{
\begin{tabular}{l|cc}
\toprule
Methods & LERF & 3D-OVS \\
\midrule
LERF \cite{kerr2023lerf}
& 37.4 & 54.8 \\

LEGaussians \cite{shi2024language}
& 46.6 & 53.5 \\

G-Grouping \cite{ye2024gaussian}
& 29.6 & 89.0 \\

3D-OVS \cite{liu2023weakly}
& 42.2 & 86.8 \\

OpenGaussian \cite{wu2024opengaussian}
& 38.4 & -- \\

LangSplat \cite{qin2024langsplat}
& 51.4 & 93.4 \\

Feature-3DGS \cite{zhou2024feature}
& -- & 87.8 \\

Laser \cite{miao2025laser}
& 53.0 & 89.3 \\

SAGA \cite{cen2025segment}
& -- & \underline{96.0} \\

Seg-Splat \cite{lu2025segment}
& \underline{52.1} & 91.3 \\

COS3D \cite{zhu2025cos3d}
& 50.8 & -- \\

\rowcolor{pink!60}
Ours
& \textbf{59.1} & \textbf{96.1} \\
\bottomrule
\end{tabular}}
\label{tab:lerf_ovs}
\end{table}

\begin{table}[t]
\centering
\caption{\textbf{Efficiency comparison on LERF and 3D-OVS.} Results are reported as training time (ms/iter) / inference time (ms/iter) / peak memory (MB). Lower values are better. }
\resizebox{0.95\linewidth}{!}{
\begin{tabular}{l|cc}
\toprule
\multirow{2}{*}{Methods}
& \multicolumn{2}{c}{Train / Infer / Memory } \\
& LERF & 3D-OVS \\
\cmidrule(lr){1-1}
\cmidrule(lr){2-2}
\cmidrule(lr){3-3}

G-Grouping \cite{ye2024gaussian}
& 105.85 / 252.88 / 10037.63
& 60.18 / 173.94 / 3603.59 \\

OpenGaussian \cite{wu2024opengaussian}
& 84.77 / \textbf{125.91} / \textbf{2099.80}
& 46.25 / \underline{66.84} / \textbf{1400.16} \\

LangSplat \cite{qin2024langsplat}
& 89.71 / \underline{135.39} / \underline{2546.68}
& 51.60 / \textbf{66.42} / \underline{1433.34} \\

Seg-Splat \cite{lu2025segment}
& \textbf{65.48} / 135.46 / 2744.55
& \textbf{28.66} / 71.78 / 1582.40 \\

\rowcolor{pink!60}
Ours
& \underline{75.99} / 278.56 / 3721.02
& \underline{38.98} / 108.92 / 2903.34 \\

\bottomrule
\end{tabular}}
\label{tab:efficiency_comparison}
\end{table}

\subsection{Boundary Reconstruction Optimization (BRO)}

Although $\mathcal{L}^{app}_{u,k}$ and
$\mathcal{L}^{sem}_{u,k}$ provide region-level appearance and semantic supervision, they do not explicitly emphasize object contours, where 2D-to-3D lifting errors frequently cause cross-object semantic ambiguity and geometric misalignment
\cite{zhao2019multi,cheng2021boundary}. We therefore introduce
BRO to selectively enhance reconstruction within object-boundary neighborhoods. We first render the soft object opacity map
\begin{equation}
\textstyle
A^{pred}_{u,k}(p)
=
\sum_{n=1}^{N}
\left\{
q_{n,k}\alpha_{n,u}(p)
\prod_{m<n}
\left(1-\alpha_{m,u}(p)\right)
\right\}.
\label{eq:object_opacity}
\end{equation}
Boundary bands are extracted from the predicted opacity and propagated object mask using the morphological gradient:
\begin{equation}
\textstyle
B^{pred}_{u,k}
=
\mathcal{D}(A^{pred}_{u,k})
-
\mathcal{E}(A^{pred}_{u,k}),
\,
B^{gt}_{u,k}
=
\mathcal{D}(M_{u,k})
-
\mathcal{E}(M_{u,k}),
\end{equation}
where $\mathcal{D}$ and $\mathcal{E}$ denote dilation and erosion, respectively. We combine the two boundary bands as
\begin{equation}
\textstyle
B_{u,k}
=
\operatorname{clip}
\left(
B^{pred}_{u,k}+B^{gt}_{u,k},0,1
\right)
\label{eq:boundary_union}
\end{equation}
to cover both the target contour and potentially misaligned predicted boundaries.

Let $\Delta_x$ and $\Delta_y$ denote the horizontal and vertical
forward finite-difference operators. The corresponding directional boundary weights are obtained by aligning $B_{u,k}$ with the spatial support of each finite difference and are denoted by $W^x_{u,k}$ and $W^y_{u,k}$. The BRO loss is defined as
\begin{equation}
\textstyle
\mathcal{L}^{bro}_{u,k}
=
\frac{
\displaystyle
\sum_{d\in\{x,y\}}
\left\|
W^d_{u,k}\odot
\left(
\Delta_d I^{pred}_{u,k}
-
\Delta_d I^{gt}_{u,k}
\right)
\right\|_1
}{
\displaystyle
C\sum_{d\in\{x,y\}}\|W^d_{u,k}\|_1+\epsilon
},
\label{eq:bro_loss}
\end{equation}
where $C$ is the number of image channels and $\epsilon$ ensures
numerical stability. By restricting gradient reconstruction to a
narrow contour neighborhood, BRO encourages sharper and more accurately localized object transitions without redundantly constraining homogeneous interior regions. The overall training objective is
\vspace{-0.2cm}
\begin{equation}
\textstyle
\mathcal{L}
=
\lambda_{\mathrm{app}}\mathcal{L}_{\mathrm{app}}
+
\lambda_{\mathrm{sem}}\mathcal{L}_{\mathrm{sem}}
+
\lambda_{\mathrm{bro}}\mathcal{L}_{\mathrm{bro}},
\vspace{-0.2cm}
\label{eq:overall_loss}
\end{equation}
where each loss denotes the average of its corresponding per-view, per-object objective in
\Cref{eq:appearance_loss,eq:infonce_sem,eq:bro_loss}.

\begin{figure}[t]
    \centering
    \includegraphics[width=\linewidth]{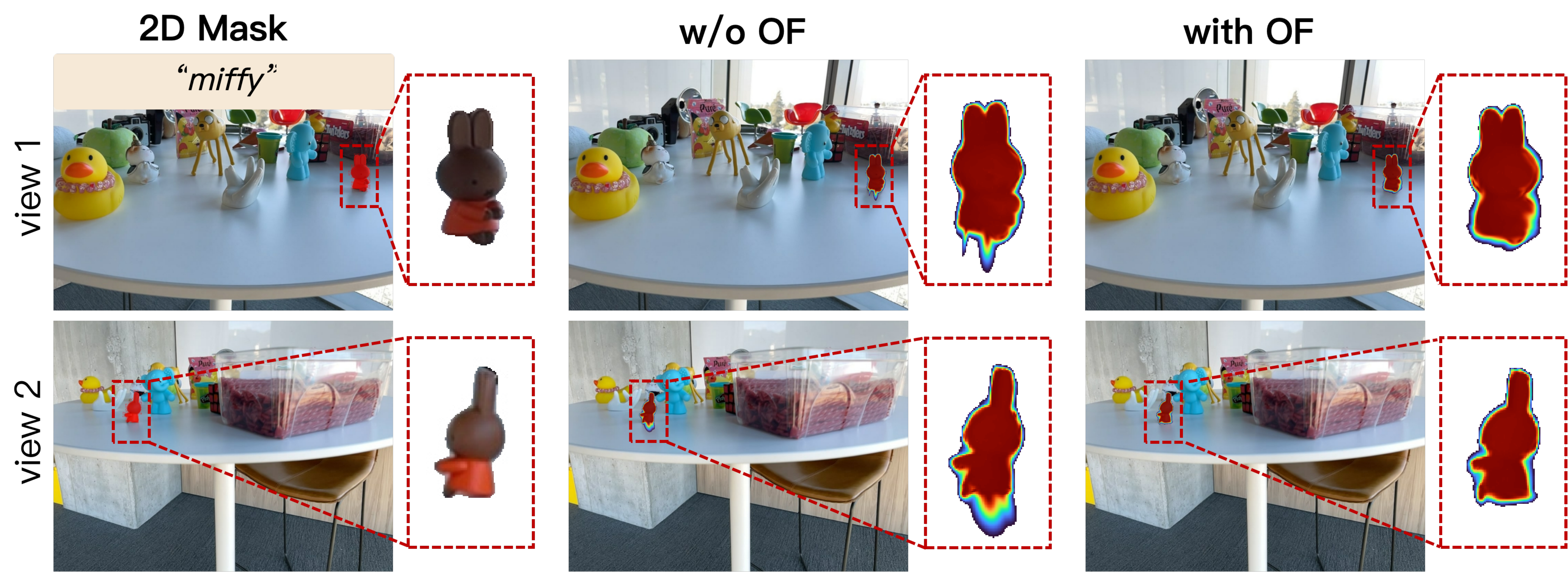}
    \vspace{-0.8cm}
    \caption{\textbf{Ablation study of OF.} We visualize the depth maps of the queried objects across multiple views.}
    \label{fig:abla_of}
    
\end{figure}

\begin{figure}[t]
    \centering
    \includegraphics[width=\linewidth]{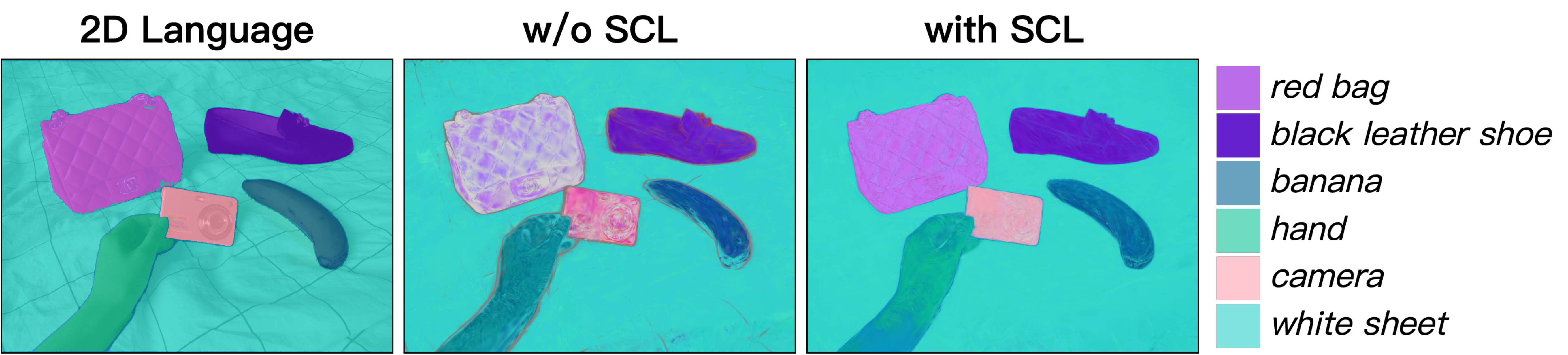}
    \vspace{-0.8cm}
    \caption{\textbf{Ablation study of SCL.} We compare the rendered 2D language maps without and with SCL against the 2D language supervision.}
    \label{fig:abla_scl}
\end{figure}

\begin{figure}[t]
    \centering
    \includegraphics[width=\linewidth]{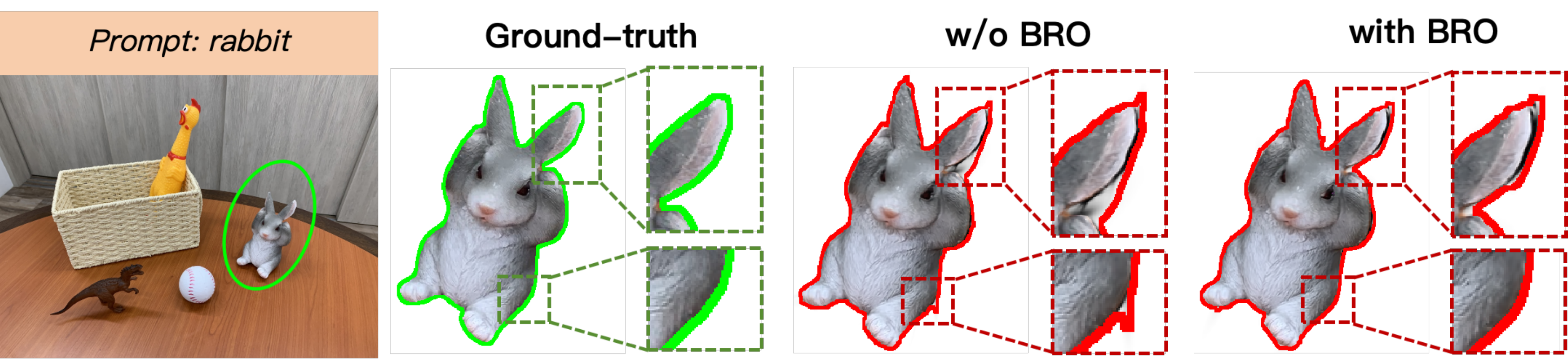}
    \vspace{-0.8cm}
    \caption{\textbf{Ablation study of BRO}. We highlight the extracted boundary bands.}
    \label{fig:abla_bro}
\end{figure}

\section{Experiments}

\subsection{Implementation Details}
We evaluate our method on LERF \cite{kerr2023lerf} and 3D-OVS
\cite{liu2023weakly}. LERF contains complex indoor scenes and emphasizes fine-grained object localization, while 3D-OVS provides language-driven annotations for indoor scenes. 2D object masks are initialized using SAM \cite{kirillov2023segment} and propagated across views using SAM2 \cite{ravi2025sam}. Image and text semantic features are extracted using OpenCLIP ViT-B/16 pretrained on LAION-2B \cite{cherti2023reproducible}.
All experiments are conducted on an NVIDIA RTX 4090 GPU.

\subsection{Results and Analysis}

As shown in \Cref{fig:qualitative_results}, our method produces more complete object regions and cleaner boundaries across different scenes and views. For the fine-grained queries ``tea in a glass'' and ``sake cup and chopsticks'', our method better preserves the queried objects while reducing interference from surrounding regions. In \Cref{tab:lerf_ovs}, our method achieves the best average mIoU on both benchmarks, reaching 59.1\% on LERF and 96.1\% on 3D-OVS. This exceeds the second-best results by 7.0 and 0.1 points, respectively.

Regarding efficiency,
\Cref{tab:efficiency_comparison} shows that our method achieves the second-fastest training speed on both datasets. Compared with OpenGaussian, the training time is reduced by 10.4\% and 15.7\%, respectively. Although our method introduces additional inference and memory costs, it offers a practical trade-off between computational overhead and segmentation performance.

\vspace{-0.2cm}
\begin{table}[t]
\centering
\caption{\textbf{Ablation study of the proposed components on LERF and 3D-OVS (\%). }}
\resizebox{0.95\linewidth}{!}{
\begin{tabular}{ccc|cc}
\toprule
\multirow{2}{*}{OF} &
\multirow{2}{*}{SCL} &
\multirow{2}{*}{BRO} &
\multicolumn{2}{c}{mAcc / mIoU / mBIoU} \\
& & & LERF & 3D-OVS \\
\cmidrule(lr){1-3}
\cmidrule(lr){4-4}
\cmidrule(lr){5-5}

-- & -- & --
& 78.01 / 52.15 / 56.52
& 97.45 / 91.30 / 80.18 \\

\checkmark & -- & --
& 79.53 / 53.03 / 56.77
& 97.95 / 92.30 / 81.99 \\

-- & \checkmark & --
& 79.59 / 53.74 / 55.45
& 97.87 / 90.41 / 80.09 \\

-- & -- & \checkmark
& 82.22 / 53.42 / 58.53
& 97.28 / 91.67 / 82.25 \\

\checkmark & \checkmark & --
& 80.04 / 56.83 / 57.23
& 98.01 / 92.68 / 82.08 \\

\checkmark & -- & \checkmark
& 81.13 / 57.56 / \underline{59.22}
& \underline{98.15} / 94.00 / \underline{85.65} \\

-- & \checkmark & \checkmark
& \underline{83.11} / \underline{57.77} / 59.14
& 98.08 / \underline{95.04} / 85.20 \\

\rowcolor{pink!60}
\checkmark & \checkmark & \checkmark
& \textbf{84.36} / \textbf{59.13} / \textbf{60.54}
& \textbf{98.59} / \textbf{96.12} / \textbf{88.27} \\

\bottomrule
\end{tabular}}
\label{tab:ablation}
\end{table}

\subsection{Ablation Studies}

\Cref{tab:ablation} evaluates the contributions of OF, SCL, and BRO on LERF and 3D-OVS. OF consistently improves performance across both benchmarks, while BRO notably enhances boundary quality. Although SCL alone yields mixed results, it provides clear complementary benefits when combined with the other components. Integrating all three modules achieves the best performance across all metrics, reaching 59.13\% and
96.12\% mIoU and 60.54\% and 88.27\% mBIoU on LERF and 3D-OVS,
respectively. Qualitative ablations in
\Cref{fig:abla_of,fig:abla_scl,fig:abla_bro} further illustrate the effects of OF, SCL, and BRO on cross-view consistency, semantic discrimination, and boundary refinement, respectively.

\section{Conclusion}

In this paper, we presented a robust open-vocabulary 3D segmentation framework based on Hybrid Gaussians with Multi-View Object Association. Hybrid Gaussians provide a unified representation for object-conditioned rendering and learning, Observation Fusion and Semantic Contrastive Learning improve identity consistency and semantic discriminability, and Boundary Reconstruction Optimization further refines contour quality and local spatial precision. 

\vfill\pagebreak

\ninept
\bibliographystyle{IEEEbib}
\bibliography{refs}

\end{document}